%% file: main.tex
\documentclass{article}

\PassOptionsToPackage{numbers, compress}{natbib}

\usepackage[preprint]{neurips_data_2024}

\usepackage[utf8]{inputenc} 
\usepackage[T1]{fontenc}    
\usepackage{hyperref}       
\usepackage{url}            
\usepackage{booktabs}       
\usepackage{amsfonts}       
\usepackage{nicefrac}       
\usepackage{microtype}      
\usepackage{xcolor}         
\usepackage{booktabs}
\usepackage{multirow}
\usepackage{color, colortbl}
\usepackage{subcaption}
\usepackage{tabu}
\usepackage{pifont}
\usepackage{makecell}
\usepackage{paralist}
\usepackage{threeparttable}
\usepackage{enumitem}
\usepackage{hhline}
\usepackage{scalerel,xparse}

\usepackage{booktabs}
\usepackage{multirow}
\usepackage{color, colortbl}
\usepackage{tabu}
\usepackage{pifont}
\usepackage{makecell}
\usepackage{paralist}
\usepackage{threeparttable}
\usepackage{enumitem}
\usepackage{hhline}
\usepackage{scalerel,xparse}
\usepackage{amsmath}

\definecolor{dt}{gray}{0.7}
\definecolor{mygray}{gray}{.9}
\definecolor{honeydew}{rgb}{0.94, 1.0, 0.94}
\definecolor{magicmint}{rgb}{0.67, 0.94, 0.82}
\definecolor{lightcyan}{rgb}{0.88, 1.0, 1.0}
\definecolor{lightgreen}{rgb}{0.86, 0.82, 1.0}
\definecolor{aliceblue}{rgb}{0.94, 0.97, 1.0}
\definecolor{babyblueeyes}{rgb}{0.63, 0.79, 0.95}
\definecolor{beaublue}{rgb}{0.74, 0.83, 0.9}
\definecolor{myblue}{rgb}{0.18, 0.33, 0.59}

\newcommand{\gray}[1]{\textcolor{gray}{#1}}

\newcommand{\sota}[1]{\colorbox{beaublue}{\textbf{#1}}}

\newcommand{\cmark}{\ding{51}}%
\newcommand{\xmark}{\ding{55}}%

\title{ VideoEval: Comprehensive Benchmark Suite for Low-Cost Evaluation of Video Foundation Model}

\author{Xinhao Li$^{1,2}$,
Zhenpeng Huang$^1$,
Jing Wang$^1$,
Kunchang Li$^2$,
Limin Wang$^{1,2}$\thanks{Corresponding author (lmwang@nju.edu.cn).}\\
$^1$State Key Laboratory for Novel Software Technology, Nanjing University \\ $^2$Shanghai AI Laboratory\\
\url{https://github.com/MCG-NJU/VideoEval}
}

\begin{document}

\maketitle

\begin{abstract}
  With the growth of high-quality data and advancement in visual pre-training paradigms, Video Foundation Models (VFMs) have made significant progress recently, demonstrating their remarkable performance on traditional video understanding benchmarks. However, the existing benchmarks (e.g. Kinetics) and their evaluation protocols are often limited by relatively poor diversity, high evaluation costs, and saturated performance metrics. In this paper, we build a comprehensive benchmark suite to address these issues, namely \textbf{VideoEval}. Specifically, we establish the \textbf{Vid}eo \textbf{T}ask \textbf{A}daption \textbf{B}enchmark (\textbf{VidTAB}) and the \textbf{Vid}eo \textbf{E}mbedding \textbf{B}enchmark (\textbf{VidEB}) from two perspectives: evaluating the task adaptability of VFMs under few-shot conditions and assessing their representation power by directly applying to downstream tasks. With VideoEval, we conduct a large-scale study on 20 popular open-source vision foundation models. Our study reveals some insightful findings on VFMs: 1) overall, current VFMs exhibit weak generalization across diverse tasks, 2) increasing video data, whether labeled or weakly-labeled video-text pairs, does not necessarily improve task performance, 3) the effectiveness of some pre-training paradigms may not be fully validated in previous benchmarks, and 4) combining different pre-training paradigms can help improve the generalization capabilities. We believe this study serves as an important complement to the current evaluation for VFMs and offers valuable insights for the future research.
\end{abstract}

\input{introduction}

\section{Related Work}

\paragraph{Video foundation models (VFMs).} 
With the continuous growth of image~\cite{cc3m,cc12m,laion} and video data~\cite{bain2021frozen,internvid,valor,vast,panda70m} and advancements in pre-training paradigms, research on Video Foundation Models (VFMs) has progressed rapidly. Current VFMs are primarily built with different pre-training paradigms: contrastive or predictive self-supervised learning methods~\cite{moco,videomoco,dino,oquab2023dinov2,vjepa} on unimodal video data, masked video modeling based on unimodal video data~\cite{st_mae,videomae,videomaev2,bevt,mvd,ominimae,hmae} and video-text contrastive learning based on multimodal visual-text pairs~\cite{cpd,videoclip,allinone,videococa,Cheng2022VindLUAR,internvid}. Some works~\cite{internvideo,umt,internvideo2,zhao2024videoprism} combine these paradigms, enabling VFMs to extend further into multimodal understanding. Additionally, some studies introduce modalities like audio and speech on top of video and text~\cite{valor,vast,internvideo2}, further expanding the capabilities of VFMs. Recently, InternVideo2~\cite{internvideo2} combines pre-training paradigms and large-scale high-quality data to scale VFMs to 6 billion parameters, achieving remarkable performance improvements.

\paragraph{Evaluation of VFMs.} 
Previous works primarily utilize action recognition benchmarks~\cite{kay2017kinetics,goyal2017something,ava} to evaluate VFMs with a focus on understanding the appearnace and motion from videos. To enhance evaluation diversity, some studies explore richer domains and tasks~\cite{howsense,deng2023bear,robustaction}, but they still remain limited to the action recognition tasks. The InternVideo series~\cite{internvideo,internvideo2} and VideoGLUE~\cite{yuan2023videoglue} attempt to provide a more comprehensive evaluation on VFMs by reporting performance on a number of benchmarks and evaluation protocols. However, these efforts are still based on existing benchmarks and incurred high validation costs. In contrast, our VideoEval considers the characteristics and application scenarios of VFMs, offering a comprehensive and more challenging evaluation solution from the perspectives of task definition and evaluation protocols.

\input{methodology}
\input{experiments}

\section{Conclusions}

We have presented the VideoEval, a comprehensive benchmark suite for efficiently evaluating the VFMs. To this end, we first establish the VidTAB benchmark, which explores suitable evaluation tasks and protocols for VFMs from the perspective of assessing their adaptability to unknown tasks with limited samples. Additionally, we create the VidEB benchmark to evaluate the capability of VFMs as direct feature embedding extractors for the downstream tasks. Based on the VideoEval, we conduct a large-scale study involving 20 popular open-source vision foundation models, providing valuable insights for future research directions.

    \bibliographystyle{unsrt}
    \bibliography{main}

\clearpage
\appendix

\input{appendix}

\end{document}

%% file: introduction.tex
\section{Introduction}

\begin{figure}[t]
    \centering
    \includegraphics[width=1\linewidth]{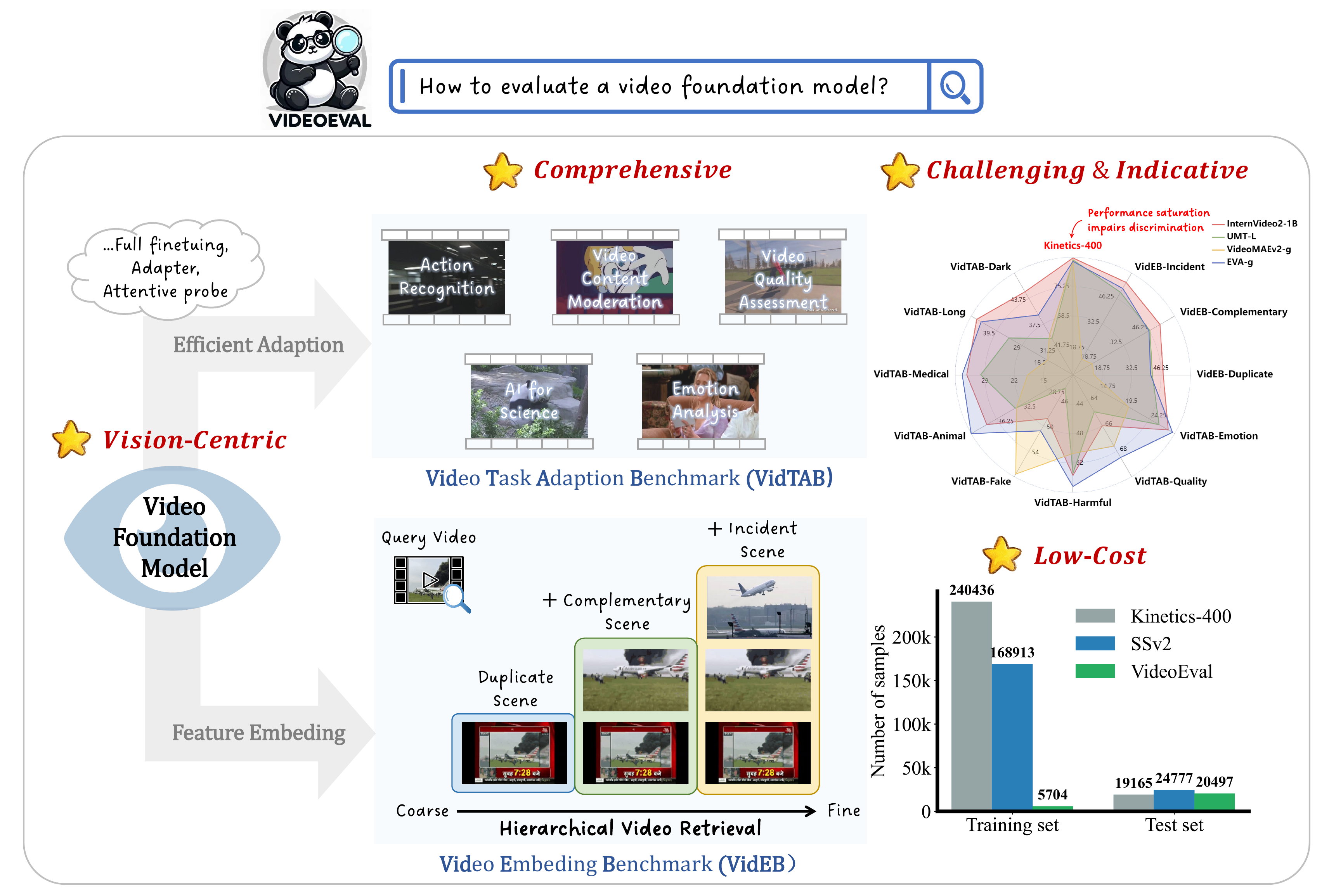}
    \caption{ \textbf{Overview of VideoEval.} We propose a novel, vision-centric evaluation method for video foundation models that is comprehensive, challenging, indicative, and low-cost. }
    \label{fig:framework}
     \vspace{-2mm}
\end{figure}

The field of deep learning is experiencing a paradigm shift of pre-training and transfer learning due to the emergence of foundation models (FMs). These models, exemplified by BERT~\cite{devlin2018bert}, GPT~\cite{gpt3, gpt4, gpt4v}, CLIP~\cite{clip} and Stable Diffusion~\cite{stablediffusion}, are pre-trained on massive and diverse data at scale and demonstrate remarkable generalization power to a broad spectrum of downstream tasks.

In the realm of video understanding, early researchers train backbone networks~\cite{tsn,slowfast,tdn,timesformer,videoswin,mvit} using visual classification tasks on large-scale labeled datasets like ImageNet~\cite{imagenet} and Kinetics~\cite{kay2017kinetics}. However, the high cost associated with labeled data promotes the development of self-supervised learning methods that capitalize on unlabeled data for visual pre-training~\cite{videomoco,maskfeat,st_mae,videomae,bevt}. Furthermore, researchers delve into multimodal pre-training utilizing large-scale weakly-supervised visual-text pairs~\cite{cpd,videoclip,videococa,internvid,umt}, thereby enhancing their models' capabilities and demonstrating impressive zero-shot performance. Overall, fueled by the accumulation of high-quality image and video data and advancements in visual pre-training paradigms, Video Foundation Models (VFMs) witness remarkable progress in recent years. A new generation of VFMs~\cite{st_mae,videomae,videomaev2,vjepa,zhao2024videoprism,internvideo,internvideo2} emerges, demonstrating outstanding performance on conventional video understanding benchmarks.

The rapid development of VFMs raises a new problem: \textbf{\textit{How to evaluate a video foundation model?}} In image realm, previous works assess the generalization capability of Image Foundation Models (IFMs) by evaluating their performance on numerous downstream visual tasks, encompassing diverse scenarios and evaluation protocols~\cite{vtab,imagenet_nae,imagenetv2,imagenet_rob,robust2,imagenet_x,battle}. In contrast, previous works primarily evaluates VFMs through benchmarks simply focusing on action recognition tasks~\cite{videomae,vjepa,yuan2023videoglue}. Some studies~\cite{mvbench,internvideo2,zhao2024videoprism} have also considered combining VFMs with Large Language Models (LLMs) to evaluate their performance on multimodal video understanding tasks. There are several issues with current evaluation benchmarks: \textbf{(1)} {\bf Limited domain and task}: benchmarks like Kinetics~\cite{kay2017kinetics}, Something-Something~\cite{sth} and AVA~\cite{ava}, which focus on action recognition, overlook other video understanding scenarios (e.g., video quality assessment), limiting their applicability in evaluating the generalization capabilities of visual foundation backbones across diverse video understanding applications. \textbf{(2)} {\bf Saturated performance}: the performance of VFMs on the standard benchmarks~\cite{k400} has reached a saturation point ( around 90\% Top-1 accuracy), making it challenging to differentiate between the true capabilities of different VFMs. \textbf{(3)} {\bf  High evaluation cost}: the conventional validation protocols often necessitate end-to-end training on the entire dataset, resulting in a significant challenge for large VFMs. \textbf{(4)} {\bf Less indicative}: Incorporating LLM into VFMs for evaluation on multi-modal video tasks may introduce bias and is less indicative for vision representation power, as performance differences might stem from the language model rather than the VFMs itself. 

To tackle these problems, we build a comprehensive vision-centric benchmark suite for evaluating VFMs, namely VideoEval. As shown in Figure \ref{fig:framework}, our method has the following key features: \textbf{\textit{Comprehensive}}: First, we create the Video Task Adaptation Benchmark (VidTAB) to evaluate the adaptability of VFMs to unseen tasks with limited training samples. We collect public datasets from various video task domains, ranging from action recognition in special scenarios, AI for science, video content moderation, video quality/aesthetic assess, to emotion analysis. From these domains, we construct eight adaptation tasks and propose evaluation protocols and adaptation methods suitable for current VFMs. Additionally, to assess the capability of VFMs as a direct feature embedding extractor for downstream applications, we create the Video Embedding Benchmark (VidEB), which includes four tasks to evaluate their embedding power at different granularities.
\textbf{\textit{Challenging \& Indicative}}: Due to the diversity of test data and the more challenging evaluation protocols, our VideoEval can effectively distinguish between various VFMs that perform similarly on traditional benchmarks, providing deeper insights and more accurate evaluation on their true capabilities.
\textbf{\textit{Low-cost}}: Thanks to our training-light few-shot evaluation and training-free feature embedding evaluation protocols, VideoEval requires significantly fewer evaluation cost compared to previous benchmarks, while still maintaining a comparable number of testing samples to ensure accurate and stable evaluations.
\textbf{\textit{Vision-centric}}: Our evaluation focuses solely on the capabilities of video foundation models themselves, avoiding the bias due to the introduction of LLMs.

Based on VideoEval, we evaluate 20 open-source vision foundation models, including Video Foundation Models (VFMs), Image Foundation Models (IFMs), and IFMs with image-to-video adaptors. \textbf{Our main findings as following:} First, current VFMs still struggle to adapt to unseen video tasks with limited training samples. Second, while more data and larger models generally improve performance, augmenting video training data can sometimes negatively affect certain tasks. Third, the effectiveness of certain pre-training paradigms, such as VideoMAEv2~\cite{videomaev2}, may not have been adequately validated in previous benchmarks. Finally, combining multiple pre-training paradigms can lead to models with better generalization capabilities, such as performing multimodal contrastive learning after unimodal visual self-supervised pre-training~\cite{umt,internvideo2}.

\input{tables/compare_bench}

%% file: tables/compare_bench.tex
\begin{table}[t]
\centering
\caption{\textbf{Comparison of VFMs Benchmark.} "Num. training" denotes number of training samples, "Num. test" denotes number of test samples, and "Beyond Action" denotes the tasks in this benchmark extend beyond action understanding.}
\label{tab:vfms_benchmark}
\resizebox{\textwidth}{!}{
    \begin{tabular}{c|cc|cccc}
        \toprule
        \textbf{Benchmark} & \textbf{Num. training} & \textbf{Num. test}  & \textbf{Beyond Action} & \textbf{Task Diversity} & \textbf{Domain Diversity} & \textbf{VFMs-specific protocol} \\
        \midrule
        \rowcolor{mygray}\multicolumn{7}{c}{\textit{Single-dataset Benchmarks}}\\
        \midrule
        Kinetics-400~\cite{k400} & 240,436 & 19,165  & \xmark & \xmark & \xmark & \xmark \\
        Sth-Sth V2~\cite{goyal2017something}  & 168,913 & 24,777   & \xmark & \xmark & \xmark & \xmark \\
        Moment-in-Time~\cite{mit}  & 791,246 & 33,898   & \xmark & \xmark & \xmark & \xmark \\
        UCF101~\cite{ucf101}   & 9,537 & 3,783  & \xmark & \xmark & \xmark & \xmark\\
        
        \midrule
        \rowcolor{mygray}\multicolumn{7}{c}{\textit{Multi-dataset Benchmarks}}\\
        \midrule
        SEVERE~\cite{thoker2022severe} & 868,446  & 144,830  & \xmark & \cmark & \cmark & \xmark \\
        BEAR~\cite{deng2023bear}  & 240,236 & 140,436  & \xmark & \cmark & \cmark & \xmark \\
        VideoGLUE~\cite{yuan2023videoglue}    & 1,896,621 & 239,011   & \xmark & \cmark & \cmark & \cmark \\
        \midrule
        \textbf{VideoEval} & 5,704 & 20,497  & \cmark & \cmark & \cmark & \cmark \\
        \bottomrule
    \end{tabular}%
}
\end{table}

%% file: methodology.tex
\section{Building VideoEval}

\begin{figure}[t]
    \centering
    \includegraphics[width=1\linewidth]{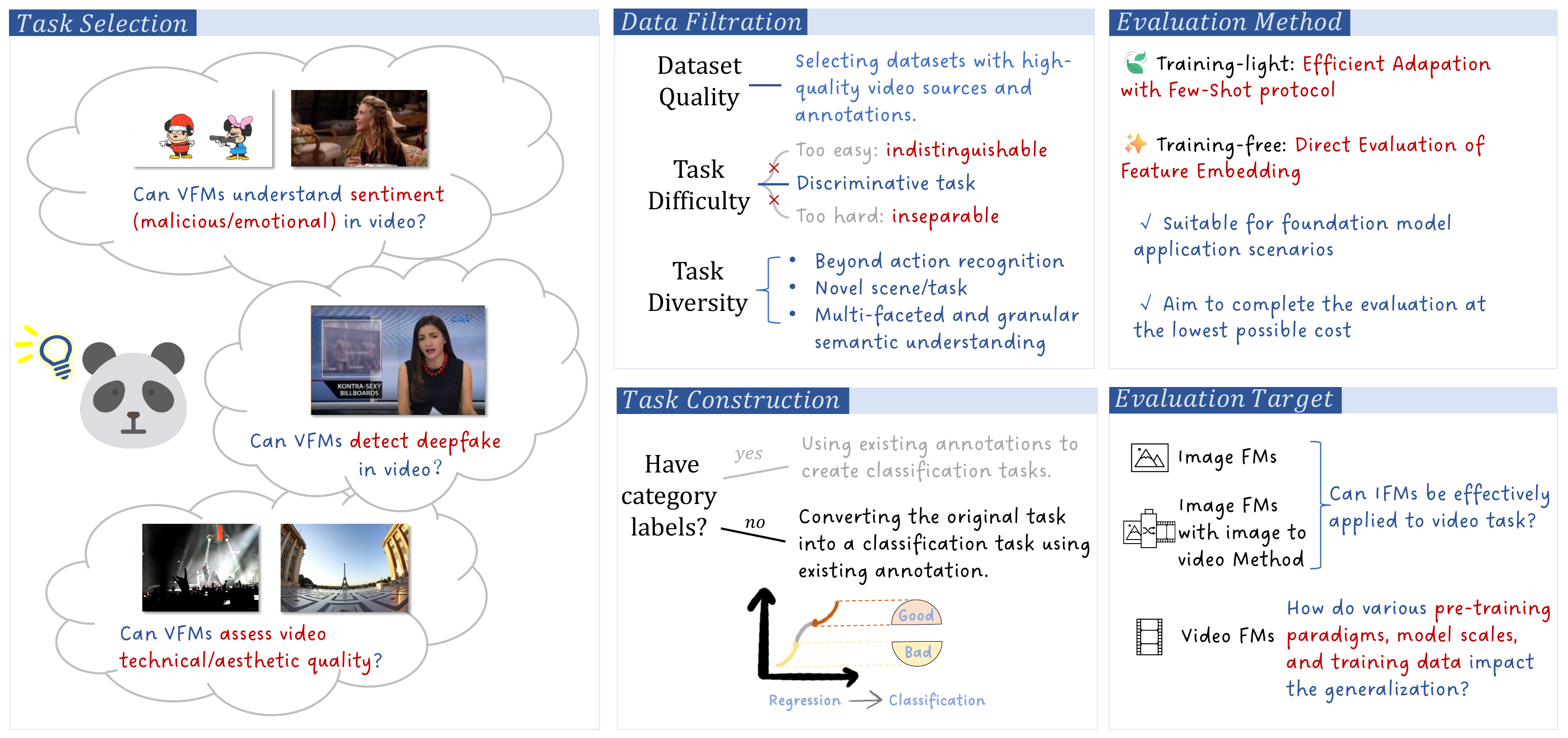}
    \caption{ \textbf{Illustration of building VideoEval. }We build VideoEval through the following steps: (1) conducting task selection by considering our expected capabilities for VFMs. (2) performing data filtration from the perspectives of quality, difficulty, and diversity, (3) standardizing the task format through task construction, and (4) defining evaluation methods and targets. }
    \label{fig:build_videoeval}
\end{figure}

We argue that a powerful video foundation model should possess two key capabilities: (1) strong task adaptation ability, i.e., the ability to \textbf{\textit{adapt to diverse and new tasks with limited training samples}}, and (2) the capacity of general video encoder to \textbf{\textit{extract feature embedding that retain and distill key information from videos}}, directly supporting various downstream tasks. From these two perspectives, we construct VideoEval, which includes the Video Task Adaptation Benchmark (VidTAB) and the Video Embedding Benchmark (VidEB). By collecting diverse tasks from different scenarios and employing low-cost evaluation scheme, our VideoEval can efficiently and comprehensively assess the generalization ability of VFMs in video understanding. In this section, we present our VideoEval in detail. The construction pipeline for VideoEval is illustrated in Figure \ref{fig:build_videoeval}, and the evaluation tasks we ultimately constructed are presented in Table \ref{tab:task_dimension}.

\subsection{Video Task Adaption Benchmark}

\paragraph{Collecting diverse datasets from public source.} Previous benchmarks primarily focus on evaluating video models for action recognition, overlooking many other video understanding tasks. Therefore, we consider five different application scenarios: \textit{Action Recognition in Special Scenarios} (\textbf{Action}), \textit{AI for Science} (\textbf{Science}), \textit{Video Content Moderation} (\textbf{Safety}), \textit{Video Quality Assessment} (\textbf{Quality}), and \textit{Emotion Analysis} (\textbf{Emotion}). We collect publicly annotated video datasets from the Internet. Subsequently, we manually filter out data with low video or annotation quality. We also consider the difficulty level of tasks to ensure that our benchmark can distinguish between models. We discard tasks that are very easy or can even be accomplished without video understanding, as well as tasks that are extremely difficult even for humans. It does not imply that we have low expectations for foundation model capabilities relative to humans; Rather, we aim to ensure meaningful and effective evaluation in the context of task adaptation, because overly challenging tasks would not provide clear distinctions between models. 

\input{tables/task_details}

\paragraph{Constructing the adaptation tasks based on the existing annotations.} Classification tasks are straightforward and well-defined, with high classification performance often indicating effective feature learning. Therefore, they are suitable for evaluating video foundation models. We construct adaptation classification tasks based on the collected data and annotations. For datasets that originally include category labels, such as ARID~\cite{xu2021arid} and Animal Kingdom~\cite{animal}, we select categories with sufficient samples to ensure evaluation accuracy and stability. We also control the final number of categories to avoid making the adaptation task overly difficult. For datasets that are not originally classification tasks, we transform the tasks into classification tasks. For example, for DOVER~\cite{dover}, which is used for video aesthetics and technical quality assessment (a regression task), we assume that videos with quality scores in the top 40\% are "high-quality videos" and those with scores in the bottom 40\% are "low-quality videos", thus converting the original task into a binary classification task. In total, we construct eight classification tasks to evaluate the adaptation capabilities of video foundation models. Specific construction details of each task are provided in the Appendix.

\begin{figure}[t]
    \centering
    \includegraphics[width=0.95\linewidth]{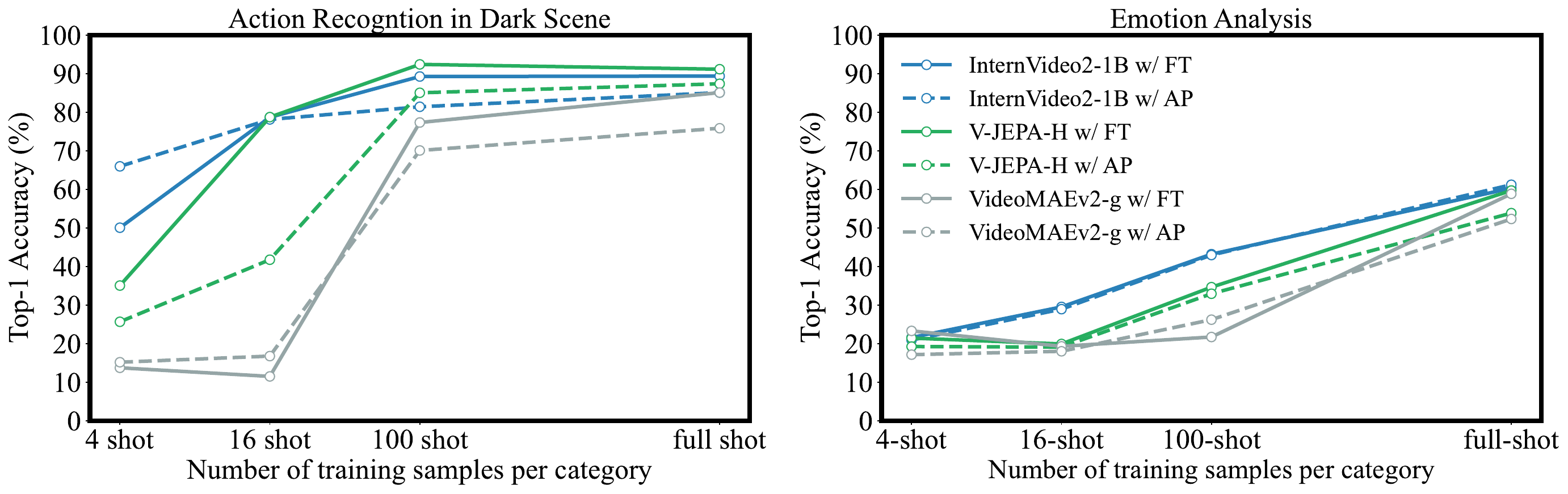}
    \caption{\textbf{Performance comparison on different training data scales.} We evaluate the performance variation of multiple video foundation models across tasks from two different domains as the scale of the training data changed. 'FT' and 'AP' denote full finetuning and attentive probe, respectively.}
    \label{fig:lowshot}
\end{figure}

\paragraph{Determining the evaluation protocol and metric.} Previous studies~\cite{internvideo,internvideo2,yuan2023videoglue} typically fine-tune video foundation models using the entire samples of training set, and most popular benchmarks have large training sample sizes. We argue that this evaluation method overlooks the examination of the adaptation capability of VFMs under few-shot setting. As illustrated in Figure~\ref{fig:lowshot}, under the scenario of using full training samples, the differences between VFMs are difficult to discern. However, under a low-sample protocol, different foundation models exhibit varying degrees of task adaptation capabilities. We observe that for the task of Action Recognition in Dark Scenes, which VFMs usually excel at, there are significant differences in adaptation capabilities among different models when training samples are extremely limited (4 shot and 16 shot). As the number of samples gradually increases to 100 shot, these differences diminish. Conversely, for more challenging tasks like Emotion Analysis, the performances of different models are uniformly weak when training samples are extremely limited, showing no discernible differences until a certain number of training samples (100 shot) are reached, at which point different models begin to demonstrate distinct adaptation capabilities. Therefore, to account for the adaptation capabilities of models with different numbers of training samples, we define a new task adaptation capability evaluation score (TA-score):
\begin{equation}
    \text{TA-score} =  \frac{{Acc}^{4s}+Acc^{16s}+Acc^{100s}}{3},
\end{equation}
where \(Acc^{4s}\), \(Acc^{16s}\), \(Acc^{100s}\) represent the model's top-1 accuracy for 4-shot, 16-shot, and 100-shot classifications, respectively. Unless otherwise specified, we will use the TA-score as the metric to evaluate the performance of various tasks in VidTAB.

\begin{figure}[t]
    \centering
    \includegraphics[width=0.85\linewidth]{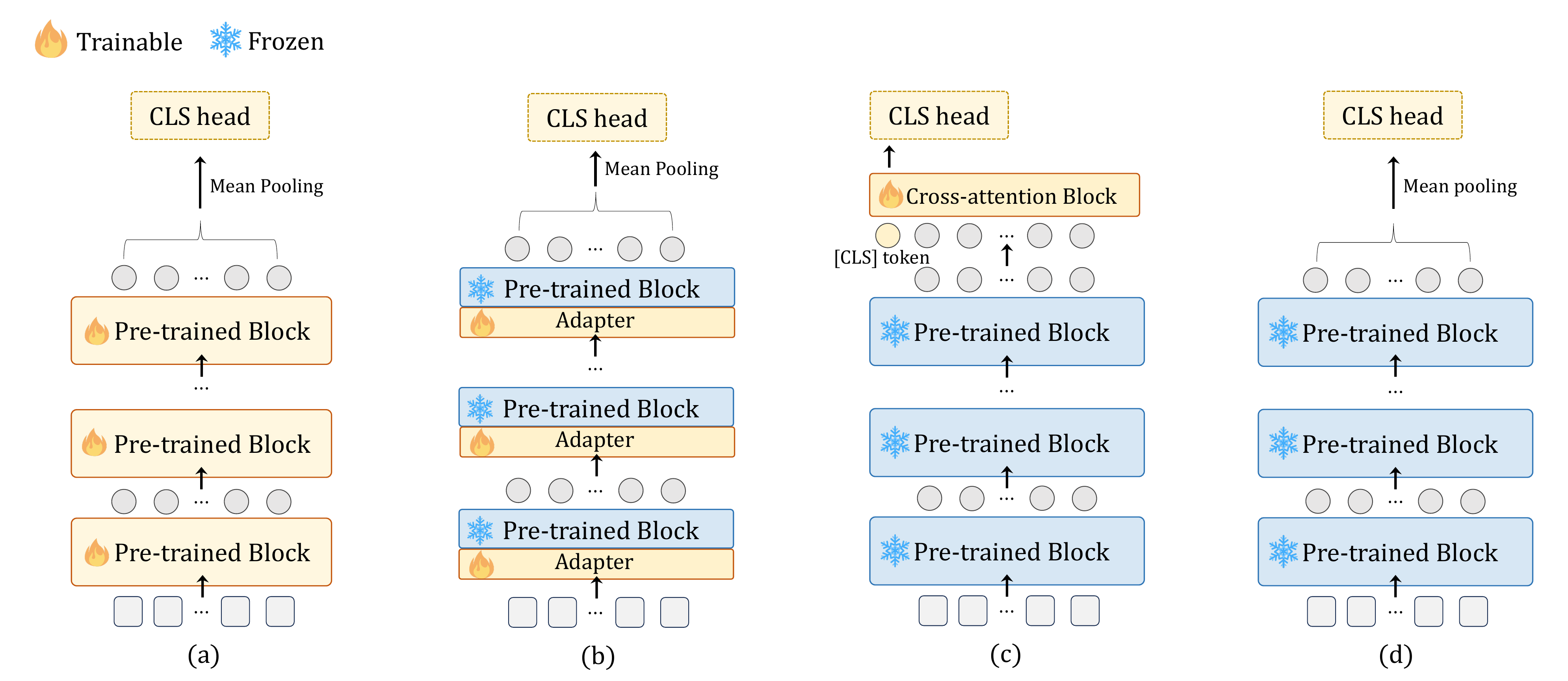}
    \caption{ \textbf{Illustration of different adaptation method}: (a) Full Finetuning, (b) Adapter,  (c) Attentive Probe, and (d) Linear Probe.}
    \label{fig:adaptation_method}
\end{figure}
\input{tables/adapation_method}

\paragraph{Identifying efficient adaptation method for evaluation.} We also need to investigate how to adapt the video foundation models to the downstream tasks. Previous work~\cite{peft,visualtuning,vpt,provpt,st_adapter,yangaim,li2023zeroi2v} has explored various strategies for efficient adapting the foundation models. Here, we consider several popular methods, as illustrated in Figure \ref{fig:adaptation_method}: \textbf{Full Finetuning}: Fine-tuning all the parameters of the pre-trained model. \textbf{Adapter}: Freezing the pre-trained model and inserting learnable low-rank adapter~\cite{adapterhub} modules into each block of the pre-trained model for adaptation. \textbf{Attentive Probe}: Freezing the pre-trained model and adding an additional learnable cross-attention block at the end of the model to achieve attentive pooling, followed by a linear projection for classification. \textbf{Linear Probe}: Directly using the features from the pre-trained model, performing mean pooling, and then using a linear projection for classification. We evaluate the performance of these adaptation methods based on the V-JEPA-H model, as shown in Table \ref{tab:speed}. Full finetuning and adapter achieve the best adaptation performance, but require high training costs. Linear probe is highly efficient but showed weak adaptation performance. Attentive probe offers a good trade-off between efficiency and adaptation performance. Therefore, in subsequent evaluation experiments, we employ attentive probe to adapt various video foundation models.

\subsection{Video Embedding Benchmark}

Apart from the traditional classification tasks, the more direct evaluation of representations typically involves the retrieval tasks such as video action retrieval~\cite{action_retrieval1, action_retrieval2, action_retrieval3}, which primarily rely on text information. However, these benchmarks often overlook the overall scene context and exhibit an overlap with recognition tasks. In contrast, inspired by previous works~\cite{fickr,dvsc23,ccweb,vcdb, DISC}, we focus on vision-centric embedding evaluation in Table \ref{tab:task_dimension}. Specifically, we require models to rank and recall instance-level samples based on similarity, rather than solely on action categories. This evaluation protocol focuses on the model's capability of capturing the subtle visual information.

\textbf{Evaluation protocol.} To facilitate fine-grained embedding evaluation, we include two tasks for assessment: \textbf{(1) Hierarchical Video Retrieval} aims to retrieve videos from a database that closely matches the query video in terms of scene, viewpoint, and temporal context. According to previous work \cite{fivr5k}, videos related to the query are categorized into three levels based on their similarity to the query: Duplicate Scene Videos (DSVs), Complementary Scene Videos (CSVs), and Incident Scene Videos (ISVs), as shown in Table \ref{tab:task_dimension}.
Consequently, the retrieval tasks are structured into three hierarchical levels:
\begin{itemize}
    \item Duplicate Scene Video Retrieval: only DSVs are positive instances.
    \item Complementary Scene Video Retrieval: both DSVs and CSVs are positive instances.
    \item Incident Scene Video Retrieval: DSVs, CSVs, and ISVs are all positive instances.
\end{itemize}
For the evaluation metric, we follow \cite{fivr5k} to utilize the mean Average Precision (mAP) to assess the quality of video ranking. 
\textbf{(2) Video Copy Detection} aims to detect edited copies of the query video. In contrast to the ranking/retrieval task where all video pairs need to be sorted according to video embedding similarity, it is required to identify a set of video pairs that contain edited versions of the given query. Following \cite{dvsc23}, we consider the micro-AP ($\mu$AP) as our evaluation metric that operates on all queries jointly and takes the confidence scores
into account. 

%% file: tables/task_details.tex
\begin{table*}[tp]
    \caption{
    \textbf{Task details of VideoEval.} All videos are collected from the public datasets for building tasks of VidTAB and VidEB.}
    
    \label{tab:task_dimension}
    \centering
    \setlength\tabcolsep{4pt}
    \resizebox{1.0\textwidth}{!}{
        \begin{tabular}{c|c|c|l}
        \Xhline{1.0pt}
        \textbf{Domain} & \textbf{Task} & \textbf{Source} & \textbf{Task Description} \\
        \Xhline{1.0pt}

    \multicolumn{4}{l}{\gray{\textit{\textbf{Video Task Adaptation Benchmark (VidTAB)}}}} \\
        \hline
        
        \multirow{4}{*}{\textbf{\makecell[c]{Action Recognition\\in Special Scenarios}}} & Action Recognition & \multirow{2}{*}{ARID~\cite{xu2021arid}} & \cellcolor{gray!5}{\textit{\textcolor{myblue}{Recognizing 11 distinct human actions in dark scenarios.}}} \\
        ~ & in Dark Scene & ~ & \cellcolor{gray!5}{e.g. Run / Walk / Drink } \\
        \hhline{~|-|-|-}
        ~ & Action Recognition & \multirow{2}{*}{BreakFast~\cite{breakfast}} & \cellcolor{gray!5}{\textit{\textcolor{myblue}{Classifying 10 types of long-duration cooking videos.}}} \\
        ~ & in Long Video & ~ & \cellcolor{gray!5}{e.g. Milk / Tea / Sandwich} \\
    
        \hline
        \multirow{3}{*}{\textbf{\makecell[c]{AI for Science}}} & Medical Surgery & {SurgicalActions160~\cite{surgicalactions160}} & \cellcolor{gray!5}{\textit{\textcolor{myblue}{Classifying 16 surgical actions in gynecologic laparoscopy.}}    e.g. Knotting / Suction / Injection} \\
        \hhline{~|-|-|-}
        ~ &\multirow{2}{*}{\makecell[c]{Animal Behavior}}&\multirow{2}{*}{\makecell[c]{Animal\\Kingdom~\cite{animal}}}& \cellcolor{gray!5}{\textit{\textcolor{myblue}{Classifying 12 behaviors of wild animals from diverse environmental footage.}}} \\
        ~ & ~ & ~ & \cellcolor{gray!5}{e.g. Flying / Chirping / Preening} \\
        
        \hline
        \multirow{4}{*}{\textbf{\makecell[c]{Video Content\\Moderation}}} &  \multirow{2}{*}{\makecell[c]{ Fake Face }}& \multirow{2}{*}{\makecell[c]{FaceForensics++~\cite{faceforensics++}}} & \cellcolor{gray!5}{\textit{\textcolor{myblue}{Determine whether the faces in the video have been tampered with by AI technology (such as DeepFake).}}} \\
        ~ & ~ & ~ & \cellcolor{gray!5}{e.g. Origin video / Video with fake face} \\
        \hhline{~|-|-|-}
        ~ & \multirow{2}{*}{\makecell[c]{Harmful Content}} & \multirow{2}{*}{mob~\cite{mob}} & \cellcolor{gray!5}{\textit{\textcolor{myblue}{Detecting 3 degrees of malicious content within videos.}}} \\
        ~ & ~ & ~ & \cellcolor{gray!5}{e.g. Obscene / Indecent activity / Violent activity } \\

        \hline
        \multirow{2}{*}{\textbf{{\makecell[c]{Video Quality\\Assessment}}}} & \multirow{2}{*}{\makecell[c]{Quality Assess}} & \multirow{2}{*}{DOVER~\cite{dover}} & \cellcolor{gray!5}{\textit{\textcolor{myblue}{Evaluating videos from an aesthetic and technical perspective and categorizing them into low and high quality.}}} \\
        ~ & ~ & ~ & \cellcolor{gray!5}{e.g. Low quality / High quality} \\

        \hline
        \multirow{2}{*}{\textbf{{\makecell[c]{Emotion Analysis}}}} & \multirow{2}{*}{\makecell[c]{Emotion Analysis}} & \multirow{2}{*}{CAER~\cite{caer}} & \cellcolor{gray!5}{\textit{\textcolor{myblue}{Classifying 7 different human emotions in video.}}} \\
        ~ & ~ & ~ & \cellcolor{gray!5}{e.g. Happy / Fear / Anger} \\
        \hline
         \multicolumn{4}{l}{\gray{\textit{\textbf{Video Embedding Benchmark (VidEB)}}}} \\
        \hline
        \multirow{8}{*}{\textbf{\makecell[c]{Scene Understanding\\in  Temporal Contexts}}} & \multirow{2}{*}{\makecell[c]{Duplicate Scene\\ Retrieval}}  & \multirow{2}{*}{FIVR5K~\cite{fivr5k}} & \cellcolor{gray!5}{\textit{\textcolor{myblue}{Retrieve Duplicate Scene Videos (DSV):}}} \\
        ~ & ~ & ~ & \cellcolor{gray!5}{Videos captured by the same camera and sharing at least one scene (without considering any application transformations).} \\
        \hhline{~|-|-|-}
        ~ & \multirow{2}{*}{\makecell[c]{Complementary Scene\\Retrieval}}  & \multirow{2}{*}{FIVR5K~\cite{fivr5k}} & \cellcolor{gray!5}{\textit{\textcolor{myblue}{Retrieve Complementary Scene Videos (CSV): }}} \\
        ~ & ~ & ~ & \cellcolor{gray!5}{Retrieve a portion of the same spatiotemporal segment captured from different perspectives.} \\
        \hhline{~|-|-|-}
        ~ & \multirow{2}{*}{\makecell[c]{Incident Scene\\Retrieval}}  & \multirow{2}{*}{FIVR5K~\cite{fivr5k}} & \cellcolor{gray!5}{\textit{\textcolor{myblue}{Retrieving Incident Scene Videos (ISV):}}} \\
        ~ & ~ & ~ & \cellcolor{gray!5}{The same event is close in both space and time, but there are no overlapping videos.} \\
        \hhline{~|-|-|-}
        ~ & \multirow{2}{*}{\makecell[c]{Copy Detection}}  & \multirow{2}{*}{DVSC23~\cite{dvsc23}} & \cellcolor{gray!5}{\textit{\textcolor{myblue}{Detecting edited versions of the same source video.}}} \\
        ~ & ~ & ~ & \cellcolor{gray!5}{Given a query inserted with one or more copied segments, detect the source video from the database.} \\


        \Xhline{1.0pt}
        \end{tabular}
    }
\end{table*}

%% file: tables/adapation_method.tex
\begin{table}[ht]
    \centering
    \caption{\textbf{ Comparison of adaptation method on V-JEPA-H~\cite{vjepa}} All results are obtained
     using A100-80G with PyTorch-builtin mixed precision, using a batch size of 4 to measure CUDA memory and training time. "Dark" and "Emotion" denote the tasks of Action Recognition in Dark Scenes and Emotion Analysis, respectively.}
     \resizebox{0.85\linewidth}{!}{
    \begin{tabular}{c|ccc|cc}
        \Xhline{1.0pt}
        \begin{tabular}{c}Adaptation\\method\end{tabular}  & \begin{tabular}{c}Tunable\\Params (M)\end{tabular}  & \begin{tabular}{c}CUDA\\Memory (G)\end{tabular} & \begin{tabular}{c}Training\\ Time (h)\end{tabular}  & \begin{tabular}{c}Dark\\TA-score\end{tabular}   & \begin{tabular}{c}Emotion\\TA-score\end{tabular} \\
        \Xhline{1.0pt}
          full finetuning & 663.7 & 52.1 & 1.0 & 68.8 & 25.3 \\
          adapter & 52.6 & 45.0 & 1.0 & 62.4 & 24.7 \\
          attentive probe & 19.7 & 6.4 & 0.4 & 54.7 & 23.8 \\
          linear probe & 0.0 & 6.0 & 0.3 & 12.9 & 16.2 \\

       \Xhline{1.0pt}
    \end{tabular}
    }
    \vspace{-4mm}
    \label{tab:speed}
\end{table}

%% file: experiments.tex
\section{Benchmarking Video Foundation Models}

\subsection{Video foundation models and evaluation details}

\paragraph{Evaluation targets.} 
We evaluate twenty open-source vision foundation models. Including: (1) twelve video foundation models, covering \textbf{\textit{different pre-training paradigms, model scales, and training data scales}}, to analyze the impact of these factors on the generalization capability of foundation models. (2) five image foundation models to observe \textbf{\textit{how much generalization capability trained on image data can exhibit in video understanding}}. (3) three image-to-video methods based on image foundation models to assess the \textbf{\textit{effectiveness of current efficient transfer methods}}.

\paragraph{Implementation details.} 
All models take 8 frames (16 frames if the model has temporal downsampling), with each frame being 224x224 in size as input. For VidTAB, to ensure fair comparison and efficient assessment, we train all models for the same number of epochs and make minor adjustments to the hyperparameters to ensure convergence.  
For VidEB, all models take 16 frames, with each frame of size 224x224. 
In hierarchical video retrieval, the similarity of video-level embedding determines the ranking of retrieval results.
In video copy detection, each sample is segmented into 5 clips. The detection confidence score for the entire video is derived from the maximum frame-wise similarity computed for each query-reference pair. 
See the appendix for more details.

\subsection{Results on VidTAB}

\input{tables/vidtab_result}

\paragraph{Zero-shot evaluation.} 
To preliminarily test the difficulty of the dataset, we first evaluate the zero-shot performance of the eight tasks using two image language models and two video language models. As shown in the top section of Table 3, both image and video models demonstrate relatively high performance for action-related tasks, and video models achieve better. For tasks involving low-level information understanding, such as Quality Assessment task, image models perform significantly better. In contrast, for other tasks involving scenarios typically unseen in training data, such as medical surgery videos or Safety Review tasks requiring complex semantic reasoning, all models perform similar to random guess.

\paragraph{Main results of adapting foundation models.}  Table \ref{tab:vidtab_result} presents the evaluation results on VidTAB. We summarize our findings as follows.

\textbf{In general}, \textbf{(1)} Despite exhibiting a degree of generalization capability, \textit{current vision FMs still struggle to adapt to unseen video tasks with limited training samples.} VFMs outperform IFMs, particularly in tasks related to action and behavior understanding. However, IFMs achieve superior performance on more low-level tasks, particularly in the domains of safety and quality, especially when combined with image-to-video adaptation techniques. 

\textbf{(2)} The \textit{adaptation performance of models generally increases with the growth of data and model size}, as observed by the improvements observed from V-JEPA-L to V-JEPA-H (+1.5) and ViCLIP-L-10M to ViCLIP-L-200M (+1.3). 

\textbf{For the training data,} \textbf{(3)} \textit{While augmenting video training data is generally beneficial, it can negatively impact the performance on some tasks.} For both VideoMAEv2-g and InternVideo2-1B\(_{stage1}\), fine-tuning on Kinetics-710 data significantly enhances Action-related tasks, but consistently degrades the performance on the Safety and Quality tasks. Similar findings are observed with ViCLIP-L, where post-pretraining on a large-scale video dataset improves Action-related tasks but diminishes performance in other domains (Science, Safety, Quality, Emotion). It could be attributed to the limited diversity of the current video training data.
    
\textbf{(4)} For models trained on single-modal visual data, \textit{incorporating additional weak-supervised post-pretraining with visual-text data leads to significant improvements in adaptation capabilities}. This is evident in the performance gains observed from UMT-L\(_{stage1}\) to UMT-L\(_{stage2}\) (+3.6) and from InternVideo2-1B\(_{stage1}\) to InternVideo2-1B\(_{stage2}\) (+8.0). Interestingly, this finding contradicts previous conclusions drawn from commonly used action recognition benchmarks, suggesting that these benchmarks may introduce bias.

\textbf{For the pre-training paradigms of model,} \textbf{(5)} \textit{The effectiveness of pre-training paradigms in scaling model size might not be adequately validated on popular action recognition benchmarks~\cite{k400,goyal2017something}.} While VideoMAEv2~\cite{videomaev2} successfully scaled a model to 1B parameters using the dual masking strategy, its adaptation performance (37.7 vs 44.4) significantly declined compared to VideoMAEv1-H. Interestingly, VideoMAEv2-g demonstrated remarkable performance after fine-tuning on Kinetics-710 (0.66M), suggesting that the abundant labeled data may have compensated for the shortcomings of its pre-training performance. 
  
\textbf{(6)} \textit{Single-modal self-supervised pre-training paradigms exhibit superior data efficiency compared to multimodal weakly-supervised pre-training paradigms}. Notably, V-JEPA and VideoMAEv1, trained solely on relatively small-scale unlabeled video data via self-supervised pre-training, demonstrate comparable or even superior performance to ViCLIP, which is trained on a massive dataset of video-text pairs.

\textbf{Finally,} \textbf{(7)} \textit{Effective adaptation method for image-to-video FMs is crucial}. Three image-to-video methods based on CLIP-L achieved significant performance improvements compared to using an attentive probe directly. We believe this represents a promising avenue for future research.

\input{tables/Videb_result}

\subsection{Results on VidEB}
The main results of VidEB are presented in Table \ref{tab:videb_result}. We evaluate the embedding performance using different pre-training paradigms for IFMs and VFMs as frozen feature extractors. Surprisingly, \textbf{IFMs performs better than most VFMs}, likely due to the existing gap in spatial modeling capabilities between VFMs and IFMs.

\textbf{For the pre-training paradigms of the model,} 
\textbf{(1)} \textit{The contrastive learning (CL) based approach consistently excels in embedding evaluation.} 
Due to CL's emphasis on the relationships between samples during training, DINOv2, which focuses solely on vision, outperforms vision-language contrastive methods like CLIP across multiple tasks. 

\textbf{(2)} \textit{The effectiveness of masked video modeling is closely tied to the targets it reconstructs or aligns with. }
With higher semantic richness, it shows progressive improvements in embedding quality for VideoMAE-L, V-JEPA-L, and UMT-L\(_{stage1}\). 

\textbf{(3)} \textit{Vision-centric pretraining outperforms multi-modal pretraining in vision-centric scenarios.} 
Comparing UMT-L\(_{stage1}\) and InternVideo2-1B\(_{stage1}\) with their multi-modal counterparts UMT-L\(_{stage2}\) and InternVideo2-1B\(_{stage2}\), the introduction of visual-text pair data in multi-stage training does not enhance performance in vision-centric scenarios. This is also consistent with the performance differences observed between DINO and CLIP-style pre-training methods.

Additionally, we assess the \textbf{impact of fine-tuning on the embedding evaluation of these pre-trained models}. \textbf{(4)} \textit{Labels bring new semantic information or disrupt existing finer-grained semantic information.} The performance variations after fine-tuning differ based on the pre-training strategy. For UMT-L\(_{stage1}\) and InternVideo2-1B\(_{stage1}\), fine-tuning leads to a significant drop in performance (-12.1 for UMT and -16.5 for InternVideo) due to the introduction of more singular label information, which causes catastrophic forgetting. In contrast, VideoMAE and VideoMAEv2 show substantial performance gains (+14.5 and +25.8, respectively) because the low-level semantics learned during pre-training are less abstract and benefit more from the addition of high-level label information.

%% file: tables/vidtab_result.tex

\begin{table*}[t]

\caption{\textbf{Evaluating state-of-the-art FMs on the VidTAB}.  The best and second-best results of foundation models are noted by \sota{blue} and \underline{underline}, respectively. '\textbf{I}', '\textbf{V}', and '\textbf{IV}' denote image data, video data, and mixed image-video data, respectively. Data marked in \gray{gray} indicates that the model uses a model trained on that data as initialization. 'K710ft' indicates that the model was fine-tuned with supervision using the labeled action recognition dataset Kinetics-710 (0.66M).}
\begin{center}
\large
\resizebox{1\linewidth}{!}{
\begin{tabular}{lcc|c|cc|cc|cc|c|c}
\Xhline{1.0pt}
\multirow{2}{*}{} &  &  & & \multicolumn{2}{c}{\textbf{Action}} & \multicolumn{2}{c}{\textbf{Science}} & \multicolumn{2}{c}{\textbf{Safety}} & \multicolumn{1}{c}{\textbf{Quality}}  & \multicolumn{1}{c}{\textbf{ Emotion}}\\
 & \rotatebox{90}{\# Params (M)} & \rotatebox{90}{\# Pretrain Data} & \rotatebox{90}{\textbf{Average}} & \rotatebox{90}{Dark Scene}  & \rotatebox{90}{Long Video} & \rotatebox{90}{Medical Surgery} & \rotatebox{90}{Animal Behavior} & \rotatebox{90}{Harmful Content} & \rotatebox{90}{Fake Face} & \rotatebox{90}{Quality Assess}  & \rotatebox{90}{Emotion Analysis} \\
\Xhline{1.0pt}
\multirow{1}{*}{Random} & - & - & 22.7 & 9.1 & 10.0 & 6.3 & 8.3 & 33.3 & 50.0 & 50.0 & 14.3  \\

\Xhline{0.8pt}
\multicolumn{12}{l}{\gray{\textit{\textbf{Zero-shot performance of visual language models}}}} \\

CLIP-L~\cite{clip} & 428 & \textbf{I}-400M & 35.7 & 29.2 & 34.6 & 12.5 & 32.9 & 42.1 & 56.3& 65.5 & 12.9  \\
EVA-CLIP-g~\cite{eva_clip} & 1365 & \textbf{I}-2B & 36.0 & 32.8 & 37.2& 9.4& 28.5 & 39.6& 52.8& 69.5& 17.9  \\
ViCLIP-L~\cite{internvid} & 428 & \gray{\textbf{I}-400M}+\textbf{V}-200M & 33.6 & 26.2& 37.5& 8.3& 29.3& 32.1& 52.2& 53.9& 29.0  \\
InternVideo2\(_{stage2}\)~\cite{internvideo2} & 1350 & \gray{\textbf{IV}-1.1M}+\textbf{IV}-25.5M & 40.6 & 37.1& 40.2& 11.5& 45.2& 59.1& 51.3& 56.1& 24.3  \\


\Xhline{0.8pt}
\multicolumn{12}{l}{\gray{\textit{\textbf{Image Foundation Model}}}} \\
CLIP-L~\cite{clip} & 316 & \textbf{I}-400M & 42.8 & 31.2& 38.0& 32.3& 36.3& 50.3& 58.5& 67.7& 28.1  \\
SigLiP-SO~\cite{zhai2023siglip} & 444 & \textbf{I}-4.11B & 42.6 & 26.9& 37.6& 36.5& 35.7& 49.4& 58.2& 68.1& 28.2  \\
EVA-g~\cite{fang2023eva} & 1035 & \textbf{I}-2B & 45.4 & 39.7& 45.9& 34.4& 40.0& 50.8& 55.2& 68.5& 28.6  \\
DINOv2-L~\cite{oquab2023dinov2} & 317 & \textbf{I}-142M & 40.7 & 37.9& 41.0& 39.6& 35.7& 34.3& 52.0& 62.2& 23.2  \\
DINOv2-g~\cite{oquab2023dinov2} & 1165 & \textbf{I}-142M & 43.0 & 35.8& 43.5& 42.7& 35.6& 44.0& 53.2& 63.7& 25.3  \\

\Xhline{1.0pt}

\Xhline{0.8pt}
\multicolumn{12}{l}{\gray{\textit{Image Foundation Model with \textbf{image-to-video adaptation method}}}} \\

ST-Adapter-CLIP-L~\cite{st_adapter} & 328 & \textbf{I}-400M & 46.9 & 43.0& 45.0& 31.2& 39.4& 49.4& \underline{64.9}& 72.3& 29.9  \\
AIM-CLIP-L~\cite{yangaim} & 328 & \textbf{I}-400M & 48.7 & 42.3& 49.9& 38.5& 38.9& 47.0& \sota{68.3}& \sota{74.6} & 30.1  \\
ZeroI2V-CLIP-L~\cite{li2023zeroi2v} & 303 & \textbf{I}-400M & 46.5 & 41.3& 46.8& 31.2& 39.3& 47.2& 64.6& 70.6& \underline{30.6}  \\

\Xhline{0.8pt}
\multicolumn{12}{l}{\gray{\textit{\textbf{Video Foundation Model}}}} \\

ViCLIP-L-10M~\cite{internvid} & 316 & \gray{\textbf{I}-400M}+\textbf{V}-10M & 41.4 & 31.3& 41.2& 30.2& 34.2& 45.3& 53.9& 68.3& 26.8  \\
ViCLIP-L-200M~\cite{internvid} & 316 & \gray{\textbf{I}-400M}+\textbf{V}-200M & 42.7 & 36.7& 43.9& 30.2& 36.8& 46.9& 54.8& 65.4& 27.2  \\
VideoMAEv1-H~\cite{videomae} & 651 & \textbf{V}-0.24M & 44.4 & 45.1& 30.3& 35.4& 37.9& 53.9& 52.0& \underline{71.3}& 28.9  \\
VideoMAEv2-g~\cite{videomaev2} & 1037 & \textbf{V}-1.35M & 37.7 & 34.9& 18.5& 18.8& 33.3& \underline{57.6}& 50.7& 67.6& 20.4  \\
VideoMAEv2-g~\cite{videomaev2} & 1037 & \textbf{V}-1.35M+K710ft & \underline{53.5} & \sota{76.1}& \underline{72.4}& \underline{50.0} & \underline{42.6}& 43.3& 56.6& 61.5& 25.1  \\
UMT-L\(_{stage1}\)~\cite{umt}& 316 & \textbf{V}-0.66M & 40.0 & 33.9& 34.4& 30.0& 33.5& 44.1& 53.4& 64.9& 26.1  \\
UMT-L\(_{stage2}\)~\cite{umt} & 316 & \gray{\textbf{V}-0.66M}+\textbf{IV}-25M & 43.6 & 32.6& 44.0& 22.9& 39.3& \sota{62.3}& 52.8& 67.7& 27.2  \\
V-JEPA-L~\cite{vjepa} & 318 & \textbf{V}-2M & 43.0 & 50.2& 34.1& 39.6& 38.5& 41.9& 51.5& 66.9& 21.0  \\
V-JEPA-H~\cite{vjepa} & 653 & \textbf{V}-2M & 44.5 & 54.7& 36.0& 35.4& 40.5& 45.1& 52.9& 67.7& 23.8  \\
InternVideo2-1B\(_{stage1}\)~\cite{internvideo2} & 1037 & \textbf{IV}-1.1M & 45.1 & 45.4& 47.7& 33.3& 37.7& 48.9& 53.7& 66.0& 27.8  \\
InternVideo2-1B\(_{stage1}\)~\cite{internvideo2} & 1037 & \textbf{IV}-1.1M+K710ft & \sota{56.4} & \underline{75.9} & \sota{77.9} & \sota{53.1}& 45.2& 47.7& 55.6& 64.7& \sota{30.9}  \\
InternVideo2-1B\(_{stage2}\)~\cite{internvideo2} & 1037 & \gray{\textbf{IV}-1.1M}+\textbf{IV}-25.5M & 53.1 & 67.6& 69.5& 38.5& \sota{48.6}& 51.2& 55.2& 64.5& 29.4  \\

\Xhline{1.0pt}
\end{tabular}
}
\end{center}


\label{tab:vidtab_result}
\end{table*}

%% file: tables/videb_result.tex

\begin{table*}[t]
    \caption{\textbf{Evaluation of State-of-the-Art Foundation Models on the VidEB Dataset.}
"K400pt" and "K400ft" denote that the model is pre-trained and fine-tuned, respectively, using the labeled action recognition dataset Kinetics-400 (0.31M). 
MCL: Multi-modal Contrastive Learning, 
SCL: Self-supervised Contrastive Learning, 
MVM: Masked Video Modeling, 
SFT: Supervised Fine-tuning. 
Other notations are consistent with those in Table \ref{tab:vidtab_result}.}
    
\begin{center}
\large
\resizebox{\linewidth}{!}{
\begin{tabular}{lcc|c|cccc}
\Xhline{1.0pt}
\multirow{3}{*}{} &  &  & & \multicolumn{4}{c}{\textbf{Scene}} 
\\
 & {Pretrain Tasks} & {\# Pretrain Data} & {\textbf{Average}} & {Duplicate}  & {Complementary} & {Incident} & {Copyright} \\

\Xhline{0.8pt}
\multicolumn{7}{l}{\gray{\textit{\textbf{Image Foundation Model}}}} \\
CLIP-L~\cite{clip} & MCL & \textbf{I}-400M & 43.0 & 41.1 & 46.4 & 52.0 & 32.3  \\
EVA-g~\cite{fang2023eva} & MCL & \textbf{I}-2B & 37.1 & 41.4 & 46.1 & 51.7 & 9.3  \\
SigLiP-SO~\cite{zhai2023siglip} & MCL & \textbf{I}-4.11B & 38.6 & 40.6 & 45.5 & 51.5 & 16.9  \\
DINOv2-L~\cite{oquab2023dinov2} & SCL & \textbf{I}-142M & {45.6} & \underline{49.0} & \underline{53.5} & {54.3} & 25.6  \\
DINOv2-g~\cite{oquab2023dinov2} & SCL & \textbf{I}-142M & \underline{48.6} & \sota{50.5} & \sota{55.1} & \sota{56.0} & 32.8  \\

\Xhline{1.0pt}

%
%

\multicolumn{7}{l}{\gray{\textit{\textbf{Video Foundation Model}}}} \\

ViCLIP-L-10M~\cite{internvid} & MVM+MCL & \gray{\textbf{I}-400M}+\textbf{V}-10M &43.7 & 40.0&	44.4&	46.9&	43.4
  \\
ViCLIP-L-200M~\cite{internvid} & MVM+MCL & \gray{\textbf{I}-400M}+\textbf{V}-200M &39.9 & 33.1&	37.2&	40.3&	\sota{49.2}
  \\

VideoMAEv1-L~\cite{videomae} & MVM & K400pt & 12.9 & 14.5 & 15.1 & 13.2 & 8.8  \\
VideoMAEv1-L-K400ft~\cite{videomae} & MVM+SFT & K400pt+ft & 27.4 & 27.6 & 30.2 & 30.3 & 21.6  \\
VideoMAEv2-g~\cite{videomaev2} & MVM & \textbf{V}-1.35M & 11.6 & 14.8 & 15.4 & 13.4 & 2.8  \\
VideoMAEv2-g-K710ft~\cite{videomaev2} & MVM+SFT & \textbf{V}-1.35M+K710ft & 37.4 & 33.8 & 37.1 & 37.1 & {41.7}  \\
UMT-L\(_{stage1}\)~\cite{umt}& MVM & \textbf{V}-0.66M & 41.1 & 42.2 & 46.6 & 49.6 & 25.7  \\
UMT-L\(_{stage1}\)-K710ft~\cite{umt}& MVM+SFT & \textbf{V}-0.66M+K710ft & 29.0 & 26.4 & 29.4 & 30.3 & 30.0  \\
UMT-L\(_{stage2}\)~\cite{umt} & MVM+MCL & \gray{\textbf{V}-0.66M}+\textbf{IV}-25M & 34.2 & 33.4 & 37.3 & 40.6 & 25.4  \\
V-JEPA-L~\cite{vjepa} & MVM & \textbf{V}-2M & 19.7 & 21.3 & 23.9 & 21.7 & 12.0  \\
V-JEPA-H~\cite{vjepa} & MVM & \textbf{V}-2M & 20.2 & 21.5 & 23.7 & 21.2 & 14.3  \\

InternVideo2-1B\(_{stage1}\)~\cite{internvideo2} & MVM & \textbf{IV}-1.1M & \sota{50.4} & {47.3} & {52.1} & \underline{54.9} & \underline{47.3}  \\
InternVideo2-1B\(_{stage1}\)-K710ft~\cite{internvideo2} & MVM+SFT & \textbf{IV}-1.1M+K710ft & 33.9 & 30.5 & 34.2 & 34.1 & 36.9   \\
InternVideo2-1B\(_{stage2}\)~\cite{internvideo2} & MVM+MCL  & \gray{\textbf{IV}-1.1M}+\textbf{IV}-25.5M & 34.6 & 32.4 & 36.8 & 39.9 & 29.3   \\

\Xhline{1.0pt}
\end{tabular}
}
\end{center}

\label{tab:videb_result}
\end{table*}

%% file: appendix.tex
\section{Appendix}
In this appendix, we provide more details of VideoEval from the following aspects:
\begin{itemize}

    \item Details of our benchmark are in \S~\ref{sec:benchmark}.
    \item Details of training and evaluation, can be found in \S~\ref{sec:training}.
    \item Ethics etatement of the datasets are in \S~\ref{sec:license}

    \item Limitations and potential negative societal impacts are in \S~\ref{sec:limitation}
    
\end{itemize}

\subsection{Details of Benchmark}
\label{sec:benchmark}

\begin{figure}[t]
    \centering
    \includegraphics[width=1\linewidth]{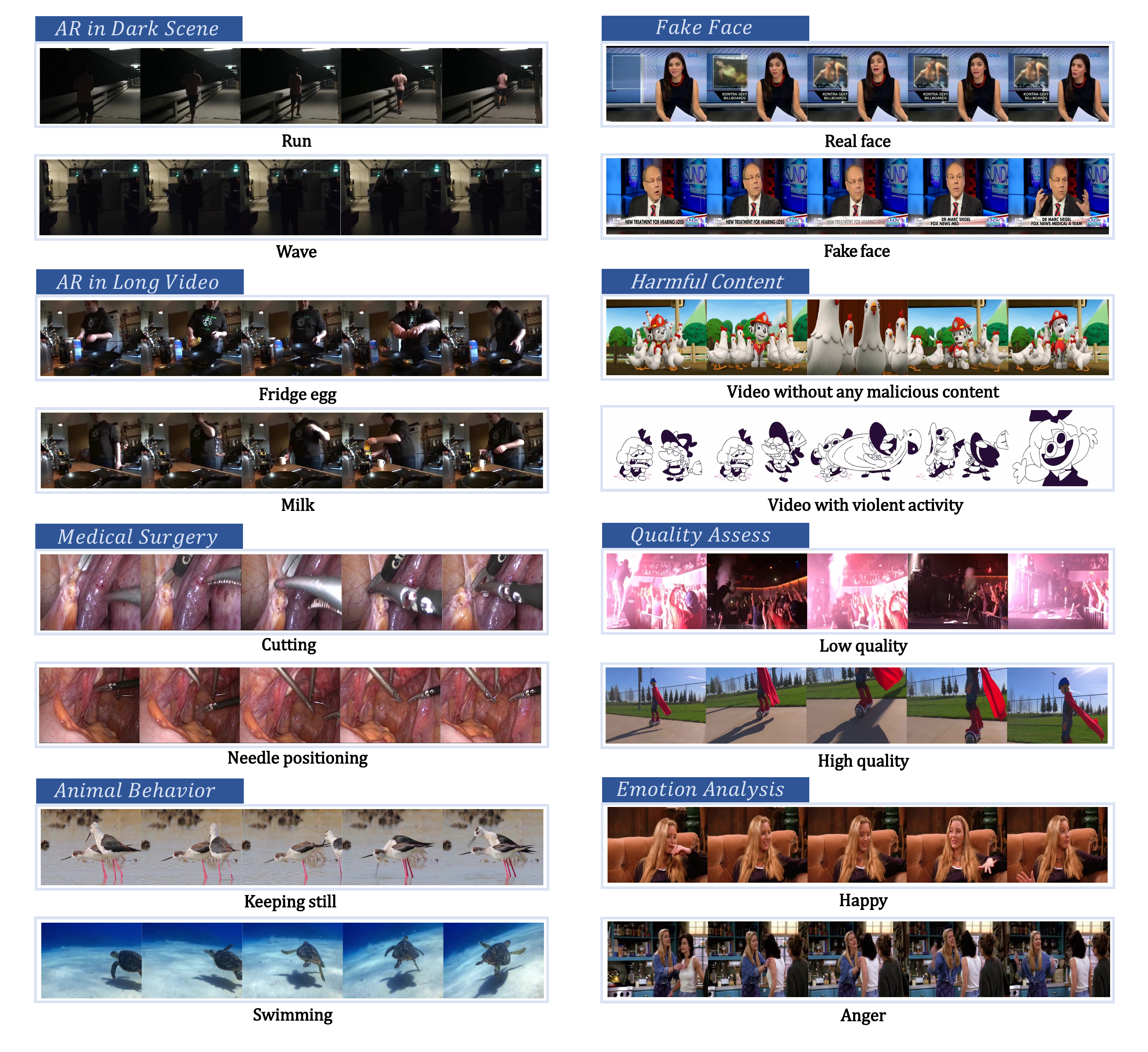}
    \caption{ \textbf{Examples of VidTAB.} We present video examples for each task in VidTAB, demonstrating that successfully completing these tasks requires VFMs to possess strong generalization capabilities.}
    \label{fig:example}
     \vspace{-5mm}
\end{figure}

\paragraph{Examples of VidTAB.} As shown in Figure \ref{fig:example}, we present some examples of tasks in VidTAB.

\input{tables/task_details_appendix}
\paragraph{Details of VidTAB.}

The detals of task construction are presented in Table \ref{tab:task_details_appendix}. For each category in one task, we sample 4, 16, and 100 samples, respectively. Given the limited volume of medical surgery data, we only sample 4 samples from each category for few-shot evaluation. To mitigate the impact of randomness, we sampled two sets of data for four tasks and obtained the benchmark results. We found that the randomness of sampling had negligible effects on the final rankings of VFMs in the benchmark.

\subsection{Details of Training and Evaluation}
\label{sec:training}

\paragraph{Code and annotations} We release the evaluation code and the modified annotation file on \url{https://github.com/leexinhao/VideoEval}, see README.md for details.

\paragraph{Checkpoints of Evaluation Models}

We provide checkpoints of the models we evaluate for reproducibility of our results.
\begin{itemize}
    \item CLIP~\cite{clip}: \url{https://huggingface.co/openai/clip-vit-large-patch14}
    \item EVA-CLIP~\cite{clip}: \url{https://huggingface.co/QuanSun/EVA-CLIP}
    \item ViCLIP~\cite{internvid}: \url{https://github.com/OpenGVLab/InternVideo/tree/main/Data/InternVid}
    \item InternVideo2~\cite{internvideo2}: \url{https://huggingface.co/collections/OpenGVLab/internvideo2-6618ccb574bd2f91410df5cd}
    \item SigLiP~\cite{zhai2023siglip}: \url{https://huggingface.co/google/siglip-so400m-patch14-384}
    \item DINOv2~\cite{oquab2023dinov2}: \url{https://huggingface.co/facebook/dinov2-giant}
    \item VideoMAE~\cite{videomae}: \url{https://github.com/MCG-NJU/VideoMAE/blob/main/MODEL_ZOO.md}
    \item VideoMAEv2~\cite{videomaev2}: \url{https://github.com/OpenGVLab/VideoMAEv2/blob/master/docs/MODEL_ZOO.md}
    \item UMT~\cite{umt}: \url{https://github.com/OpenGVLab/unmasked_teacher}
    \item V-JEPA~\cite{vjepa}: \url{https://github.com/facebookresearch/jepa}
    
\end{itemize}

\paragraph{Trainging strategies.} Specific hyperparameter configurations are available in the configs provided in our code repository. In essence, we train all models for 25 epochs using a similar training strategy, employing the Adam optimizer, a learning rate of 5e-5, and only utilizing RandomResizedCrop for data augmentation. And we use a single clip to obtain the final evaluation performance.

\paragraph{Total amount of compute and the type of resources used.} Leveraging the low cost of our evaluation protocol, we conducted each experiment involving a single VFM and a single task on one A100-80G GPU. We performed approximately 300 such experiments, each taking around 1-2 hours, resulting in a total of around 400 GPU hours.

\subsection{Ethics Statement}
\label{sec:license}

\paragraph{License of the datasets}

The dataset we are using is collected from publicly accessible sources, all licensed under Creative Commons (CC-BY) or other open-source licenses. We have diligently followed all legal requirements to integrate this data into our research, emphasizing the importance of transparency in data licensing for proper attribution and appropriate use. Although we have taken steps to ensure the inclusion of suitable content, we recognize that some problematic content may still exist. If you encounter any such content, please notify us immediately so we can take corrective action to maintain a dataset free from inappropriate material. We are dedicated to maintaining a high-quality, ethically responsible dataset and pledge to uphold principles of privacy and transparency in all our work.

\paragraph{Privacy or safety concerns in video} For personally identifiable information or offensive content in video, our data collection sources have been carefully considered, and we believe these issues are not present. However, if you discover any oversights, please do not hesitate to contact us promptly.

\subsection{Limtiations and Societal Impacts}
\label{sec:limitation}

\paragraph{Limitations.} First, due to the limitations of diversity and accuracy in our video sources and annotations, which were gathered from public resources, we plan to further enrich the task in the future by incorporating manual annotations and self-collected data. Secondly, considering the evaluation cost and simplicity, we currently only evaluate tasks like classification and retrieval, which primarily rely on VFMs' global information extraction capabilities. We have not yet considered tasks like spatio-temporal action detection and temporal grounding, which assess other aspects of VFMs' capabilities. We will expand the scope of evaluation in the future.
 
\paragraph{Potential negative societal impacts.} While our evaluation includes tasks like synthetic video recognition and harmful information recognition, these serve only as indicators of the model's overall performance in this area and cannot be used to accurately evaluate the actual performance of a specific task. If researchers or engineers in society attempt to use VFMs to perform these specific tasks, our benchmark can serve as a reference for their choice of VFMs, but it cannot be used as the final standard for evaluating that task. Otherwise, it may have negative impacts on the corresponding real-world applications.

%% file: tables/task_details_appendix.tex
\begin{table*}[tp]
    \caption{
    \textbf{Task construction details of VidTAB.} All videos are collected from the public datasets for building tasks of VidTAB.}
    
    \label{tab:task_details_appendix}
    \centering
    \setlength\tabcolsep{4pt}
    \resizebox{1.0\textwidth}{!}{
        \begin{tabular}{c|c|c|l}
        \Xhline{1.0pt}
        \textbf{Task} & \textbf{Source} & \textbf{Num. test sample} & \textbf{Details of Task Construction} \\

        \hline
        
        Action Recognition &ARID~\cite{xu2021arid} & 2011  & We directly employ the original classification task definition. Specifically, 11 categories.  \\
        \hline
        Action Recognition & BreakFast~\cite{breakfast} & 822 & We directly employ the original classification task definition. Specifically, 12 categories.  \\
    
        \hline
        Medical Surgery & {SurgicalActions160~\cite{surgicalactions160}} & 96 & We directly employ the original classification task definition. Specifically, 16 categories. \\
        \hline
        \multirow{2}{*}{\makecell[c]{Animal Behavior}}&\multirow{2}{*}{\makecell[c]{Animal Kingdom~\cite{animal}}} &\multirow{2}{*}{\makecell[c]{2268}} & Since the annotations in this dataset included multiple labels, we filtered out all categories with only single labels and then \\
         ~ & ~ & ~ & selected categories with more than 150 samples. This resulted in a final set of 12 categories. \\
        
        \hline
         \multirow{3}{*}{\makecell[c]{Fake Face}} & \multirow{3}{*}{\makecell[c]{FaceForensics++~\cite{faceforensics++}}} & \multirow{3}{*}{\makecell[c]{1800}} & We used the original 1000 videos as positive samples. Then, we divided the original videos into five parts  and used the Deepfakes, \\
        ~ & ~ & ~ & Face2Face, FaceShifter, FaceSwap, and NeuralTextures methods to generate 1000 negative samples by face-swapping. We then \\
        ~ & ~ & ~ & selected  1800 of these samples as the test set and the remaining as the training set. \\
        \hline
         \multirow{3}{*}{\makecell[c]{Harmful Content}} & \multirow{3}{*}{\makecell[c]{mob~\cite{mob}}} & \multirow{3}{*}{\makecell[c]{1661}} & We categorized videos into three classes based on their content: those containing fast repetitive movements and violence activities, \\
         ~ & ~ & ~ & those containing unpleasant appearances and obscene scenes, and those containing no malicious information at all. This resulted in a\\
         ~ & ~ & ~ & three-class classification task. \\

        \hline
        \multirow{2}{*}{\makecell[c]{Quality Assess}} &\multirow{2}{*}{\makecell[c]{DOVER~\cite{dover}}} &\multirow{2}{*}{\makecell[c]{724}} & To convert the task into a classification problem, we sorted the "overall score" label and divided the videos into positive and negative \\
        ~ & ~ & ~ & samples, with the top and bottom 40\% constituting the respective categories. \\

        \hline
        Emotion Analysis & CAER~\cite{caer} & 3953 & We directly employ the original classification task definition. Specifically, 7 categories. \\

        \Xhline{1.0pt}
        \end{tabular}
    }
\end{table*}